\documentclass[10pt,onecolumn,letterpaper]{article}
\usepackage{cvpr}
\usepackage[numbers,sort&compress]{natbib}
\usepackage{times}
\usepackage{eso-pic}

\usepackage{epsfig}
\usepackage{graphicx}
\usepackage{amsthm}
\usepackage{amsmath}
\usepackage{amssymb}
\usepackage[breaklinks=true,bookmarks=false]{hyperref}

\numberwithin{equation}{section}
\usepackage{datetime}

\usepackage[utf8]{inputenc} 
\usepackage[T1]{fontenc}    
\usepackage{hyperref}       
\usepackage{url}            
\usepackage{booktabs}       
\usepackage{amsfonts}       
\usepackage{nicefrac}       
\usepackage{microtype}      
\usepackage{xcolor}         

\usepackage{graphicx}
\usepackage{amsmath} 
\usepackage{algorithm}
\usepackage{algorithmic}
\usepackage{colortbl} 
\usepackage{tcolorbox}

\usepackage{multirow}
\usepackage{array}
\usepackage{longtable}
\usepackage{adjustbox}
\usepackage{array}
\usepackage{lscape}
\usepackage{natbib}
\usepackage[a4paper,margin=1in,headheight=15pt,headsep=24pt]{geometry}

\usepackage{fancyhdr}

\usepackage{amsmath,amsfonts,bm}

\def\eqref#1{equation~\ref{#1}}

\def\1{\bm{1}}

\DeclareMathAlphabet{\mathsfit}{\encodingdefault}{\sfdefault}{m}{sl}
\SetMathAlphabet{\mathsfit}{bold}{\encodingdefault}{\sfdefault}{bx}{n}

\usepackage{hyperref}
\usepackage{url}
\usepackage{graphicx}
\usepackage{multirow}
\usepackage{booktabs}
\usepackage{array}
\usepackage[table]{xcolor}
\usepackage{pifont}
\usepackage{subfigure}

\usepackage{algorithm}
\usepackage{algorithmic}
\usepackage{amsmath,bm}
\usepackage{amssymb}
\usepackage{amsthm}
\usepackage{multirow} 
\usepackage{graphicx}
\usepackage{amsmath}
\usepackage{amsthm}
\usepackage{setspace}
\usepackage{enumitem}

\cvprfinalcopy 
\def\cvprPaperID{****} 
\allowdisplaybreaks

\begin{document}
\lhead{}
\lfoot{\date{\today},\date{\currenttime}}
\rfoot{NGD for DL}

\title{IMAGINE: Integrating Multi-Agent System into One Model for Complex Reasoning and Planning
}
\author{
\textbf{Xikai Zhang}$^{1}$\quad
\textbf{Bo Wang}$^{3}$\quad
\textbf{Likang Xiao}$^{1}$\quad
\textbf{Yongzhi Li}$^{3}$\quad\\
\textbf{Quan Chen}$^{3}$\quad
\textbf{Wenjun Wu}$^{2,1}$\quad
\textbf{Liu Liu}$^{2,1*}$\\[0.5em]
$^1$Hangzhou International Innovation Institute, Beihang University\\
$^2$School of Artificial Intelligence, Beihang University\\
$^3$Kuaishou Technology
}
\maketitle
\begingroup
\renewcommand\thefootnote{*}
\footnotetext{Corresponding author: \texttt{liuliubh@buaa.edu.cn}}
\endgroup
\begin{abstract}
\quad \quad Although large language models (LLMs) have made significant strides across various tasks, they still face significant challenges in complex reasoning and planning. For example, even with carefully designed prompts and prior information explicitly provided, GPT-4o achieves only a 7\% Final Pass Rate on the TravelPlanner dataset in the sole-planning mode. Similarly, even in the thinking mode, Qwen3-8B-Instruct and DeepSeek-R1-671B, only achieve Final Pass Rates of 5.9\% and 40\%, respectively. Although well-organized Multi-Agent Systems (MAS) can offer improved collective reasoning, they often suffer from high reasoning costs due to multi-round internal interactions, long per-response latency, and difficulties in end-to-end training. To address these challenges, we propose a general and scalable framework called IMAGINE, short for Integrating Multi-Agent System into One Model. This framework not only integrates the reasoning and planning capabilities of MAS into a single, compact model, but also significantly surpass the capabilities of the MAS through a simple end-to-end training.
Through this pipeline, a single small-scale model is not only able to acquire the structured reasoning and planning capabilities of a well-organized MAS but can also significantly outperform it. Experimental results demonstrate that, when using Qwen3-8B-Instruct as the base model and training it with our method, the model achieves an 82.7\% Final Pass Rate on the TravelPlanner benchmark, far exceeding the 40\% of DeepSeek-R1-671B, while maintaining a much smaller model size.
\end{abstract}
\section{Introduction}
Although large language models (LLMs) have demonstrated immense potential in many areas, they still face significant challenges in complex reasoning and planning tasks. For instance, Xie et al.~\cite{xie2024travelplanner} proposed the TravelPlanner dataset, which is based on real-world travel planning scenarios. In this task, the model needs to perform complex reasoning and planning before providing a final travel plan. On this dataset, even with carefully designed prompts and prior information explicitly provided, GPT-4o achieves only a 7\% Final Pass Rate in the sole-planning mode. Similarly, even in the thinking mode, Qwen3-8B-Instruct and DeepSeek-R1-671B achieve Final Pass Rates of only 5.9\% and 40\%, respectively.

Large Language Model-based Multi-Agent Systems (LLM-MAS) have showcased remarkable proficiency in tackling intricate and multifaceted problems that demand sophisticated reasoning and planning \cite{guo2024large,zeeshan2025large,chen2024survey,qian2025scaling,liu2025symagent,qiao2025thematic,wang2025cooperative}. By harnessing the unique capabilities of multiple agents, each with distinct roles and responsibilities, these systems facilitate collaborative problem-solving through dynamic, multi-round interactions. This collective approach allows LLM-MAS to address challenges that are far more complex than those that can be effectively managed by a single-agent system, enhancing both the depth and accuracy of the solutions they generate. As such, these systems are increasingly being recognized for their potential to revolutionize fields ranging from advanced robotics to multi-agent coordination and decision-making.

Recently, a simple Reflective Multi-Agent System~\cite{shinn2023reflexion} can perform self-reflection and correction of previous reasoning, ultimately producing a more reliable response after validation. 
Chen et al.~\cite{chen2024reprompt} introduced REPROMPT, which optimizes “step-by-step instructions” in the prompts provided to LLM agents by leveraging conversation history derived from multi-agent interaction and reflection. This method achieved a Final Pass Rate of 3.89\% on TravelPlanner.
Guo et al.~\cite{guo2025mirror} proposed a Multi-Agent System that integrates intra-reflection and inter-reflection mechanisms, achieving a Final Pass Rate of 2.2\% on the TravelPlanner benchmark.
Zhang et al.~\cite{zhang2025swarmagentic} built a Multi-Agent System from scratch, inspired by Particle Swarm Optimization (PSO). This system maintains a population of candidate solutions and evolves them using feedback-driven updates, reaching a Final Pass Rate of 32.2\% on TravelPlanner.
Choi et al.~\cite{choi2025atlas} proposed a multi-agent framework called ATLAS, designed to handle the constraint-aware complexity of real-world travel planning. ATLAS incorporates a formalized methodology with dedicated mechanisms for constraint handling, iterative plan evaluation, and adaptive interleaved search.

\begin{figure}[t]
    \centering
    \includegraphics[width=0.6\linewidth]{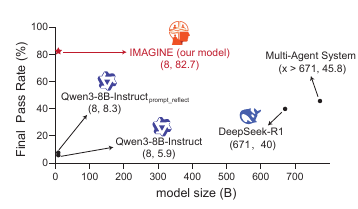}
    \caption{Final Pass Rate under different models}
    \label{fig:main-1}
\end{figure}

Despite the advantages of Multi-Agent Systems, they face several significant limitations. Firstly, these systems require the manual design of complex prompts for each agent and the construction of intricate workflows, which increases both the difficulty and cost of development. Additionally, due to the multi-turn interactions within the system, the computational overhead grows substantially as the number of pairwise interactions among agents increases, leading to significant delays in response times for individual queries. The large amount of redundant communication between agents due to multi-turn interactions further exacerbates the computational overhead. As the number of agents expands, the interaction cost tends to grow without bound, which fundamentally limits their practical deployment and application. Furthermore, the presence of multiple LLMs in the system results in high storage space consumption. If external LLM APIs are used, multi-round interactions with the APIs can incur considerable costs. Finally, the complexity of multi-agent interactions makes end-to-end training challenging, and improving performance through training becomes difficult. 

To address the aforementioned challenges, we propose a new framework called IMAGINE, which integrates a Multi-Agent System into a single model to enhance complex reasoning tasks. This framework consolidates the capabilities of an entire large-scale multi-agent system into a smaller, more efficient model, allowing a single LLM Agent to embody the abilities of a carefully designed multi-agent system. The performance is shown in Figure \ref{fig:main-1}. This approach is analogous to equipping an individual with the capabilities of a well-organized team. 
Specifically, our approach consists of three stages: New Query Generation, Multi-Agent System-based Inference Data Generation, and Agentic Reasoning Training.
In the New Query Generation stage, we augment the model’s training dataset by creating new queries, allowing the model to be exposed to more diverse training data.
In the Multi-Agent System-based Inference Data Generation stage, the newly generated queries from the previous stage are fed into a Multi-Agent System to perform inference. This process produces inference data from the Multi-Agent System, which is then used to enhance the reasoning capabilities of our model—effectively distilling the collective reasoning abilities of a well-organized and powerful team into a single, smaller model.
Finally, we propose Agentic Reasoning Training, which consists of two key components: Agentic SFT and Agentic RL. Agentic SFT takes the concatenation of the queries generated in the first stage and the corresponding Multi-Agent System-based inference data from the second stage as training data. This enables the integration of the Multi-Agent System’s reasoning capabilities into a single, much smaller model, serving as an effective cold-start stage in our approach. Agentic RL further builds upon Agentic SFT through end-to-end reinforcement learning, aiming to further strengthen the model’s agentic reasoning capabilities.

Our approach offers several key advantages. First, compared to Multi-Agent Systems, our single, compact model can be conveniently trained in an end-to-end manner to enhance performance, significantly reducing training complexity and cost. After end-to-end training, the reasoning capabilities of our model can easily surpass those of carefully designed and well-organized Multi-Agent Systems.
Moreover, since our model is a single, end-to-end compact model, it does not require the complex internal multi-turn interactions that are typical in Multi-Agent Systems. This allows for significantly reduced user wait times and greatly lowers inference costs.
In addition, being a single small model, it eliminates the need for the substantial storage space typically required by Multi-Agent Systems, as well as the cost of invoking external APIs. As a result, it offers substantial cost savings.

Our main contributions are summarized as follows:
\begin{itemize} [leftmargin=0.5cm]
	\item We propose IMAGINE, an effective, general, and scalable Agentic Reasoning Training method for a single model. This approach enables a small single model to outperform even a carefully designed and well-organized Multi-Agent System in reasoning. This is analogous to empowering an individual to surpass a well-coordinated team in reasoning ability.
    \item A scalable and general training method: The IMAGINE framework is model-agnostic. It provides a scalable and general approach to injecting Multi-Agent System behaviors into a single model, significantly improving performance through simple end-to-end training, ultimately surpassing carefully designed, well-organized Multi-Agent Systems, greatly reducing training complexity. This approach possesses broad applicability across various domains.
    \item Our method does not require the complex multi-turn interactions of a Multi-Agent System, enabling high-quality responses within a short time and significantly reducing inference costs.
	\item Using the Qwen3-8B-Instruct model as a base, we achieve an 82.7\% Final Pass Rate on the TravelPlanner dataset, greatly outperforming the DeepSeek-R1-671B model(40\%). Our model is smaller, more efficient, and delivers superior inference performance.
\end{itemize}

\begin{figure*}
    \centering
    \includegraphics[width=0.95\linewidth]{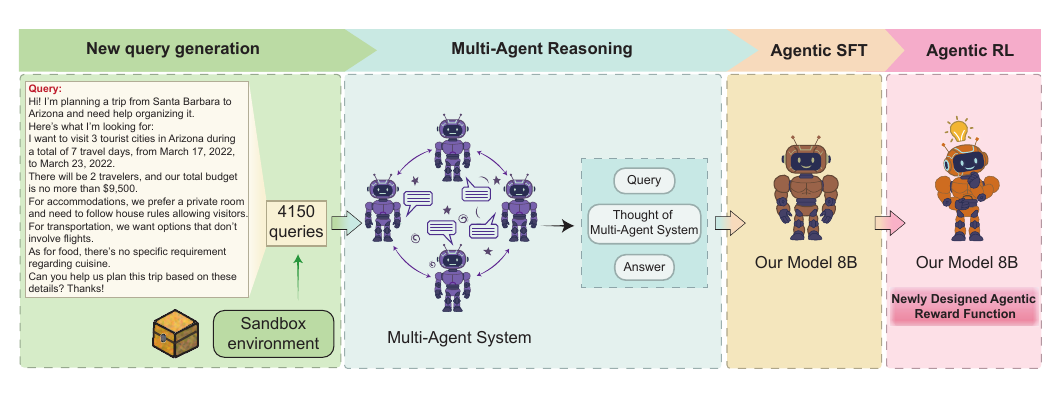}
    \caption{Overview of our proposed framework: IMAGINE.}
    \label{Overall_Flow_Chart}
\end{figure*}

\section{Related works}

\subsection{Complex Reasoning and Planning by LLMs}
Despite the remarkable progress made by large language models (LLMs) in recent years, current LLMs still struggle significantly when tackling complex reasoning and planning tasks. For instance, Xie et al. introduced a benchmark dataset called TravelPlanner~\cite{xie2024travelplanner}, which is designed to evaluate performance on realistic travel planning tasks. 
Although research by Wei et al.~\cite{wei2022chain} has demonstrated that generating Chain of Thought (i.e., sequences of intermediate reasoning steps) can substantially improve LLMs’ ability to solve complex reasoning tasks, such improvements remain quite limited: even when prompts are carefully crafted and include some prior knowledge in advance, the performance of LLMs is still far from satisfactory. For example, in the sole-planning mode of the TravelPlanner dataset, even with carefully designed prompts and the injection of prior knowledge, GPT-4o achieves only a 7\% Final Pass Rate; Even in thinking mode, Qwen3-8B-Instruct and DeepSeek-R1-671B achieve only 5.9\% and 40\% Final Pass Rates, respectively.
Subsequent work by Hao et al.~\cite{hao2024large} proposed integrating LLMs with specialized tools such as the Z3 SMT solver~\cite{de2008z3}. In this setup, LLMs can invoke the solver during travel planning, which improves performance on the TravelPlanner dataset. Although such approaches that involve the invocation of specialized tools can improve performance, this approach does not focus on enhancing the LLM's own capabilities in solving such complex reasoning problems. 

A series of subsequent works have explored Multi-Agent System based approaches. Zhang et al.~\cite{zhang2024planning} proposed a method called Planning with Multi-Constraints (PMC), a framework for LLM-based collaborative Multi-Agent System. PMC decomposes constraint-rich tasks into hierarchical sub-tasks and maps each sub-task to executable actions, thereby simplifying the planning process. 
Chen et al.~\cite{chen2024reprompt} introduced REPROMPT, which optimizes “step-by-step instructions” in the prompts provided to LLM agents by leveraging conversation history derived from multi-agent interaction and reflection. 
Guo et al.~\cite{guo2025mirror} proposed a Multi-Agent System that integrates intra-reflection and inter-reflection mechanisms, achieving a Final Pass Rate of 2.2\% on the TravelPlanner benchmark.
Yang et al.~\cite{yang2025plan} introduced Multiple Aspects of Planning (MAoP), which employs a strategist to generate planning blueprints from multiple dimensions. 
Zhang et al.~\cite{zhang2025swarmagentic} built a Multi-Agent System from scratch, inspired by Particle Swarm Optimization (PSO). 
Ou et al.~\cite{ou2025analyzing} developed an LLM-based Multi-Agent System incorporating an orchestrator agent to enhance coordination among agents. 
Choi et al.~\cite{choi2025atlas} proposed a multi-agent framework called ATLAS, designed to handle the constraint-aware complexity of real-world travel planning. 

However, all the aforementioned methods share critical limitations: such LLM-based Multi-Agent Systems face challenges such as high interaction overhead, long single-response latency, difficulties in end-to-end training, substantial resource and cost demands, and communication redundancy. These issues hinder the scalability, operational efficiency, and performance of the systems, particularly as system complexity increases and reliance on external APIs becomes more pronounced.

\subsection{LLM-based Multi-Agent Systems (LLM-MAS)}
In recent years, with the rapid development of large language models (LLMs), LLM-based Multi-Agent Systems have emerged as a research hotspot. These systems construct multiple LLM agents with distinct roles and functionalities to collaboratively accomplish complex tasks. Through multi-turn interactions, role specialization, and reflective mechanisms, LLM-based Multi-Agent Systems exhibit human-like collaboration and the potential for collective intelligence.
Recently, the Reflective LLM Agent System ~\cite{shinn2023reflexion} simulates human-like reflection mechanisms to detect and correct reasoning errors in a timely manner, thereby improving the quality of final outputs. MIRROR ~\cite{guo2025mirror} is a multi-agent system that leverages the reflective capabilities of LLMs by combining internal reflection before actions with interactive reflection after actions to enhance reasoning performance. MetaGPT ~\cite{hong2024metagpt} transforms Standard Operating Procedures (SOPs) into structured prompt chains and assigns diverse roles to different LLM agents in order to optimize workflow efficiency. CAMEL ~\cite{li2023camel}, or Communicative Agent Framework, introduces a prompting strategy known as inception prompting, which guides LLM agents to complete tasks aligned with human objectives.

Despite growing interest in LLM-MAS, critical limitations hinder their deployment and scalability. These include the need for complex manual prompt design and workflows, rapidly escalating inference costs with more agents, excessive storage and API costs, difficulty in optimizing through end-to-end training, and inefficient computation due to redundant multi-turn interactions.

\subsection{Reinforcement Learning for Reasoning Model}
The use of reinforcement learning (RL) to enhance reasoning in large language models (LLMs) has garnered significant attention~\citep{cheng2025revisiting,zhang2025critique,xiong2025minimalist}, primarily due to its ability to foster self-improvement without relying on human-annotated datasets. This is typically achieved by fine-tuning the models on complex reasoning tasks, aiming to cultivate diverse reasoning strategies~\citep{gandhi2025cognitive,yue2025does}. Key advancements, such as OpenAI o1~\citep{jaech2024openai} and DeepSeek-R1~\citep{guo2025deepseek}, demonstrate that RL techniques can be successfully deployed in large-scale commercial applications, significantly enhancing reasoning capabilities and revealing new emergent skills, such as extended reasoning chains.
In recent developments, reinforcement learning has been guided by scalar feedback signals~\citep{jaech2024openai,guo2025deepseek,liu2025understanding,yu2025dapo}. 
Among the commonly utilized algorithms in this field are online policy optimization methods, such as REINFORCE~\citep{10.1007/BF00992696}, Proximal Policy Optimization (PPO)~\citep{schulman2017proximal}, Group Relative Policy Optimization (GRPO)~\citep{shao2024deepseekmath}, and Decoupled Clip and Dynamic Sampling Policy Optimization (DAPO)~\citep{yu2025dapo}.

However, despite their success in specialized domains, research into the application of RL algorithms to more complex tasks—such as planning—remains limited. Furthermore, there is a lack of sufficient empirical evidence and well-grounded reasoning to support the adaptability and effectiveness of existing RL methods for such planning tasks.

\begin{figure}
    \centering
    \includegraphics[width=0.7\linewidth]{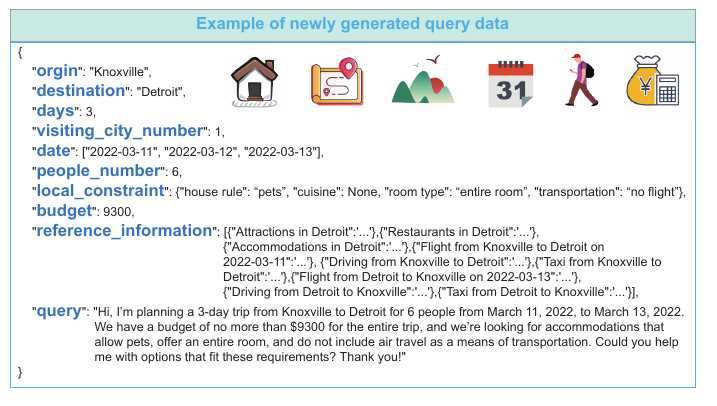}
    \caption{An example of new query. Among these, "origin" indicates the departure city, "destination" indicates the destination, "days" indicates the total number of travel days, "visiting\_city\_number" indicates the total number of cities visited during the trip, "date" indicates the specific travel date range, "people\_number" indicates the number of travelers, "local\_constraint" indicates the user's other hard requirements for this trip (i.e., hard constraints), "budget" indicates the travel budget, "reference\_information" indicates the essential reference information required for conducting travel planning, and "query" indicates the user's travel query.}
    \label{new_query_example}
\end{figure}
\section{Methods}

In this section, we use TravelPlan~\cite{xie2024travelplanner} (a complex reasoning and planning task) as an example to introduce our general and scalable method, IMAGINE. IMAGINE consists of three key stages: New Query Generation, Multi-Agent System based Inference Data Generation, and Agentic Reasoning Training. New Query Generation expands the diversity of query data in the training set by generating novel queries; Multi-Agent System based Inference Data Generation using MAS to generate inference data for queries produced in the previous stage; Agentic Reasoning comprises two components: Agentic SFT, which fine-tunes a single model using the MAS-based inference data to inject the reasoning abilities and behaviors of a MAS into a single model; Agentic RL, which further enhances the model’s agentic reasoning abilities through end-to-end reinforcement learning, ultimately enabling the single model to surpass the performance of the original MAS. Additionally, it should be noted that our research focuses on the sole-planning mode of the TravelPlanner dataset, which is used to evaluate the ability of LLMs to perform complex reasoning and planning under the condition of having all the detailed and necessary information in advance (e.g., attractions, restaurants, accommodations and transportation).

\subsection{New Query Generation}

Due to the limited size of the original TravelPlanner dataset, which contains only 1,225 queries (45 for training, 180 for validation, and 1,000 for testing), if we remove the 1,000 queries from the test set, only 225 queries remain. This small number of queries not only lacks diversity but also cannot support the training requirements of the model. Therefore, we need to generate new queries for subsequent model training. We adopted the query generation method from TravelPlanner~\cite{xie2024travelplanner} and based it entirely on the information in the sandbox environment provided by TravelPlanner. We start by focusing on basic elements such as the departure city, destination, and travel date range, and randomly combine these elements to form the framework of each query. Then, we adjust the travel duration and difficulty to create queries with different complexities.

Specifically, from the perspective of travel duration, we created new queries with three durations: 3 days, 5 days, and 7 days. A 3-day query visits one city, a 5-day query visits two cities, and a 7-day query visits three cities. From the perspective of difficulty, we created new queries with three difficulty levels: easy, medium, and hard. The difficulty level is primarily determined by the number of hard constraints—the more hard constraints there are, the higher the difficulty of the query.
Furthermore, data deduplication is crucial. That is, we must ensure that the newly generated query data does not overlap with the existing query data in the original dataset. To achieve this, we have incorporated strict deduplication logic in the code. Specifically, we check key information, such as the origin and destination, to ensure that the newly generated queries do not overlap with those in the original dataset and that there are no internal duplicates in the newly generated query data.

Finally, based on the above elements, we use GPT-4o to generate natural language queries. Using this approach, we generated 4,105 new queries in total. Example queries are shown in Figure \ref{new_query_example}. The newly generated queries ensure diversity because even small variations in the elements lead to significantly different travel plans.
It is also worth noting that, as we focus on the TravelPlanner dataset's sole-planning mode (which primarily evaluates the model's ability to reason and plan using pre-existing reference information), we also need to construct reference information for each newly generated query. These reference information include details on attractions, restaurants, accommodations, and transportation. For this, we developed code to generate corresponding reference information for each query, ensuring that this information is entirely sourced from the TravelPlanner sandbox environment.

It is important to note that, unlike the original TravelPlanner dataset, we do not guarantee that all newly generated queries will have a feasible travel plan that satisfies all constraints. Some queries may have no feasible travel plan. For example, certain cities in the reference information may lack available transportation options, making it impossible to plan a feasible trip. However, this does not affect subsequent model training, as our primary goal is for the model to learn the reasoning process. Even in the case of unsolvable queries, the model's reasoning on such problems still provides valuable training material.
%


\begin{figure*}[t]
    \centering
    \includegraphics[width=1\linewidth]{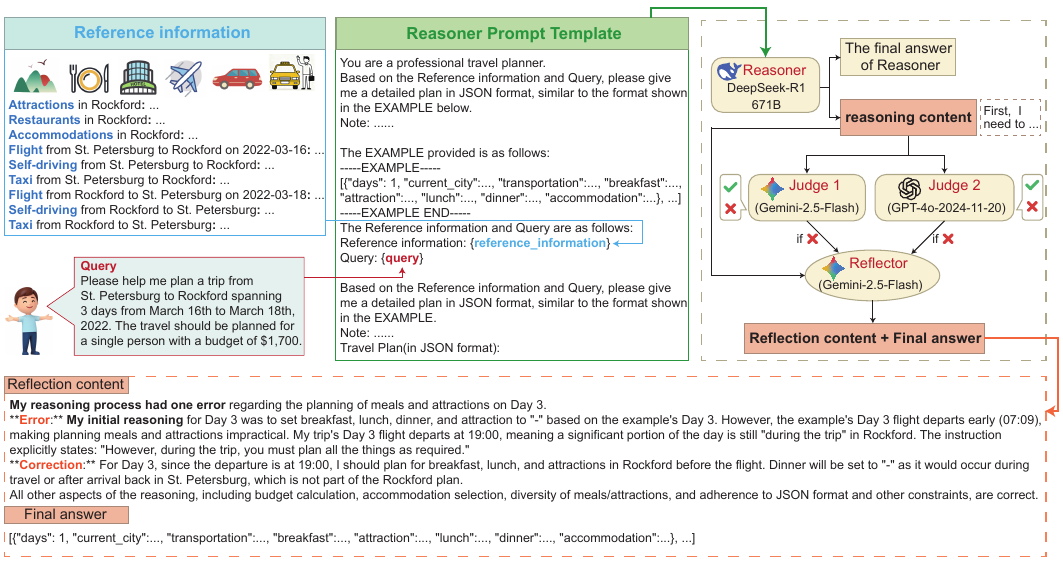}
    \caption{Inference Data Generation Process by Our Designed Multi-Agent System}
    \label{Inference Data Generation Process by Our Designed Multi-Agent System}
\end{figure*}

\subsection{Multi-Agent System based Inference Data Generation}

A standalone LLM Agent performs poorly in problems that require complex reasoning and planning, such as TravelPlanner. Research  \cite{shinn2023reflexion,guo2025mirror} has shown that reflection on complex reasoning tasks helps improve both the reasoning and the quality of the final answer. Therefore, we designed a simplified Multi-Agent System with reflection capabilities to generate high-quality reasoning data for subsequent model training, as shown in Figure \ref{Inference Data Generation Process by Our Designed Multi-Agent System}. Empirical results confirm that the reasoning data produced by our designed Multi-Agent System with reflection capabilities is of higher quality than that produced by a standalone LLM Agent (See the "Comparison of the final travel plan" section in the Experiments).

In our Multi-Agent System, we defined three different roles: Reasoner, Judge, and Reflector. The specific data generation process within the Multi-Agent System is as follows:
First, the query and reference information are placed into the Reasoner Prompt Template and input to the Reasoner (DeepSeek-R1-671B). Based on the query and reference information, the Reasoner generates the reasoning content and answer (note that this is only the answer provided by the Reasoner, and it may not be the Final answer). Then, two Judge models independently review the reasoning content generated by the Reasoner to identify any errors. The Judges only check for errors. If a Judge detects an error, they output “Errors exist.”; if no errors are found, they output “No errors.”
Whenever any Judge detects an error, the Reflector (Gemini-2.5-Flash) begins to reflect on the reasoning content generated by the Reasoner. The Reflector will first point out the errors made by the Reasoner in the reasoning content, and then provide corresponding corrections (We refer to this part of the Reflector’s output as the Reflection content). Finally, the Reflector outputs the corrected final answer (i.e., the JSON-formatted travel plan).
If both Judges detect no errors, the Reflector is not invoked, and the answer generated by the Reasoner is directly used as the Final answer.

We use the content generated by the Multi-Agent System in the above process to construct training data for subsequent Agentic Supervised Fine-Tuning. The specific construction method for the training data is as follows:
\begin{itemize} [leftmargin=0.5cm]
\item If any Judge detects an error, we construct the data in the following format:

Reasoner Prompt Template(reference information, query) +  <think>Reasoner’s reasoning content + "REFLECTION(Now, I need to reflect on whether there are any errors in my reasoning above):" + Reflector’s Reflection content + "The reflection is over, now IMMEDIATELY output the final answer!"</think> + Final answer

\item If both Judges detect no errors, we construct the data in the following format:

Reasoner Prompt Template(reference information, query) +  <think>Reasoner’s reasoning content + "REFLECTION(Now, I need to reflect on whether there are any errors in my reasoning above):" + "No errors." + "The reflection is over, now IMMEDIATELY output the final answer!"</think> + Final answer
\end{itemize}

The architecture and process are shown in Figure \ref{Inference Data Generation Process by Our Designed Multi-Agent System}.
Using this Multi-Agent System’s reasoning method, we generated intermediate reasoning data and final answers for a total of 4150 query data, including 4105 newly generated queries and the training set of 45 queries from the original TravelPlanner dataset. This data will be used for subsequent model training.


\subsection{Agentic Reasoning Training}
In this section, we provide a detailed introduction to our Agentic Reasoning Training method. Agentic Reasoning Training consists of two sequential components: Agentic Supervised Fine-Tuning (SFT) and Agentic Group Relative Policy Optimization (GRPO). Agentic SFT integrates the capabilities of a Multi-Agent System into a single model, while Agentic GRPO builds upon this foundation to conduct end-to-end training, thereby enhancing and further stimulating the model’s agentic reasoning abilities.

\subsubsection{Agentic SFT} 
Given a dataset
$\mathcal{D} = \{ (x_i, y_i) \}_{i=1}^N$,
where $x_i$ is the input (e.g., a prompt or query) and $y_i$ is the desired response, SFT optimizes the conditional likelihood of generating $y_i$ given $x_i$:
\begin{align*}
    \mathcal{L}_{\text{SFT}}(\theta)
= - \mathbb{E}_{(x,y)\sim \mathcal{D}}
\left[ \sum_{t=1}^{|y|} \log \pi_\theta (y_t \mid x, y_{<t}) \right],
\end{align*}
where $\pi_\theta$ is the model with parameters $\theta$, $y_{<t}$ denotes all tokens before time step $t$. The loss is the cross-entropy between the model’s predicted distribution and the true tokens.

We use the training data constructed in the previous stage (i.e., the Multi-Agent System based Inference Data Generation stage) to perform full SFT training on the Qwen-3-8B-Instruct model.


\subsubsection{Agentic GRPO}  
Original Group Relative Policy Optimization (GRPO)~\cite{shao2024deepseekmath} eliminates the need for a value model by evaluating the relative advantage of each response within a set of responses to the same query.
Specifically, GRPO optimizes the following objective:
\begin{align*}
\mathcal{J}_{\text{GRPO}}(\theta) 
=\mathbb{E}
\left[ \frac{1}{G} \sum_{i=1}^G \frac{1}{|y_i|} \sum_{t=1}^{|y_i|} 
\min \Big( w_{i,t}(\theta) \hat{A}_{i,t}, 
\operatorname{clip}\big(w_{i,t}(\theta), 1-\epsilon, 1+\epsilon\big) \hat{A}_{i,t} \Big) \right],
\end{align*}
where $G$ is the number of generated responses to each query $x$ (i.e., the group size), and the importance ratio $w_{i,t}(\theta)$ and advantage $\hat{A}_{i,t}$ of token $y_{i,t}$ are:
\begin{align}
w_{i,t}(\theta) 
= \frac{\pi_\theta(y_{i,t} \mid x, y_{i,<t})}{\pi_{\theta_{\text{old}}}(y_{i,t} \mid x, y_{i,<t})},
\qquad
\hat{A}_{i,t} = \frac{r(x,y_i) - \operatorname{mean}\left(\{r(x,y_j)\}_{j=1}^G\right)}
{\operatorname{std}\left(\{r(x,y_j)\}_{j=1}^G\right)},
\end{align}


\noindent respectively, where all the tokens in $y_i$ share the same advantage as $\hat{A}_i$.

\begin{figure}
    \centering
    \includegraphics[width=0.6\linewidth]{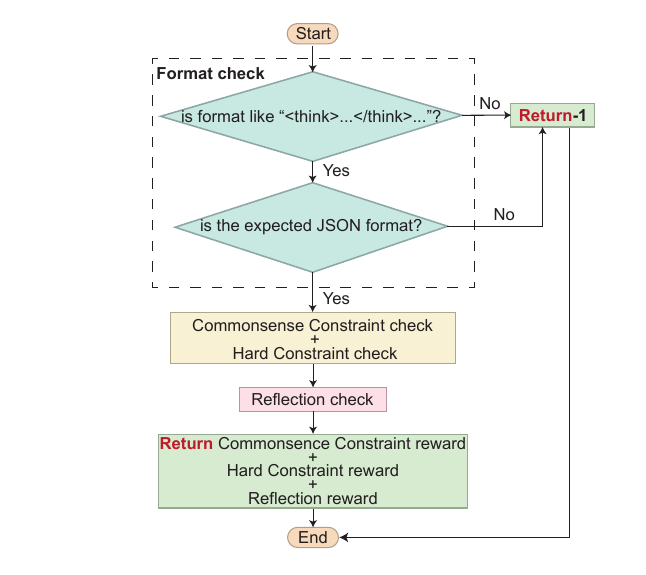}
    \caption{ Designed rule-based reward}
    \label{rule-based reward}
\end{figure}

To better suit our task, we designed a custom rule-base reward function for GRPO. Specifically, We designed the reward function to include three components: format check, constraint satisfaction check, and reflection check, as shown in Figure \ref{rule-based reward}:

\singlespacing

\noindent\textbf{Format check}:
In this part, we first strictly verify whether the model’s output follows the format "<think> ... </think> ...", i.e., it must start with "<think>", followed by some content, then "</think>", and subsequently some additional content.
If this format is not satisfied, a reward of -1 is immediately assigned and the process terminates directly. 
If it is satisfied, we further check whether the final answer output after the "</think>" tag conforms to the expected JSON format. If the format does not comply with the specified JSON format, a reward of -1 is also directly assigned and the process terminates directly. If both conditions are met, it indicates the format check has passed, and we proceed to subsequent checks.

\singlespacing

\noindent\textbf{Constraint satisfaction check}:
In this part, we perform two checks: \textit{commonsense constraints check} and \textit{hard constraints check}. Specifically, the \textit{commonsense constraints check} include: "is reasonable visiting city", "is valid restaurant", 
"is valid attraction", "is valid accommodation", "is valid transportation", "is valid information in the current city", "is valid information in the sandbox", and "is not absent".
The \textit{hard constraints check} include: "is valid cuisine", "is valid room rule", "is valid transportation", "is valid room type", and "is valid cost".
Using the rule-based evaluation code provided by TravelPlanner, we can determine whether each of the above items is satisfied. For both commonsense and hard constraints, we assign rewards based on the proportion of satisfied items to the total number of items. Specifically:
\begin{itemize}
    \item \texttt{commonsense constraint reward }= number of satisfied commonsense constraints items / total commonsense constraints items.
    \item \texttt{hard constraint reward }= number of satisfied hard constraints items / total hard constraints items.
\end{itemize}
Finally, we sum the \texttt{commonsense constraint reward} and \texttt{hard constraint reward} to obtain the total \texttt{constraint satisfaction reward}.

\singlespacing

\noindent\textbf{Reflection check}:
In this part, we check whether the model has included a reflection at the end of the "<think>" section. Specifically, we use a regular expression to verify this. If the model has included a reflection at the end of the "<think>" section, the \texttt{reflection reward} will be set to +0.5. If not, the \texttt{reflection reward} will be set to -0.5.

\noindent In summary, the reward function we have designed to suit our task is as follows:
\begin{align}
R=\left\{\begin{matrix}
 -1 ,&\text{if the format check does not pass} 
 \\
 \hat{R},&\text{if the format check passes} 
\end{matrix}\right.
\end{align}
where $\hat{R}$ is the sum of 

\texttt{commonsense constraint reward}, \texttt{hard constraint reward}, 
and \texttt{reflection reward}.

\newcommand{\uparrowtext}[1]{\textcolor{green!30!black}{\raisebox{0.1em}{\fontsize{7pt}{7pt}\selectfont ↑}\raisebox{0.05em}{\fontsize{7pt}{7pt}\selectfont#1}}}
\newcommand{\downarrowtext}[1]{\textcolor{red}{\raisebox{0.1em}{\fontsize{7pt}{7pt}\selectfont ↓}\raisebox{0.05em}{\fontsize{7pt}{7pt}\selectfont#1}}}
\definecolor{sigmaBG}{HTML}{D9FFDD}
\definecolor{baselineBG}{HTML}{EDEDED}
\definecolor{SequentialBG}{HTML}{D3F3FE}

\section{Experiments}
\subsection{Experimental Setup}
\noindent\textbf{Datasets.}
During the Agentic SFT stage, we utilize the newly generated 4,105 queries (an example of which is shown in Figure \ref{new_query_example}), along with the training set (45 queries) from the original TravelPlanner \cite{xie2024travelplanner} dataset, to construct the training set for the Agentic SFT stage with the method described in the section "Multi-Agent System based Inference Data Generation";
In the Agentic RL stage, we use the training set (45 queries) and the validation set (180 queries) from the original TravelPlanner dataset, totaling 225 queries, as the training set for the Agentic RL stage;
During testing, we consistently select the test set (1,000 queries in total) from the original TravelPlanner dataset as our test set. For testing, we first have the model under evaluation perform inference on the test set, and then use the rule-based evaluation code provided in TravelPlanner to assess the model's inference results.

 \singlespacing

\noindent\textbf{Criteria.}
We use the following six evaluation criteria from \cite{xie2024travelplanner} to assess the performance of different models:
1) \textit{Delivery Rate}: The ratio of successfully generating the final travel plan (Regardless of the quality of the travel plan, as long as it can be successfully generated). 2) \textit{Commonsense Constraint Micro Pass Rate}: The ratio of the number of commonsense constraints passed to the total number of commonsense constraints. 3) \textit{Commonsense Constraint Macro Pass Rate}: The proportion of solutions that completely pass all commonsense constraints among all test solutions. 4) \textit{Hard Constraint Micro Pass Rate}: The ratio of the number of hard constraints passed to the total number of hard constraints.5) \textit{Hard Constraint Macro Pass Rate}: The proportion of solutions that completely pass all hard constraints among all test solutions. 6) \textit{Final Pass Rate}: The ratio of successfully generating the final travel plan that satisfies all constraint conditions. This metric is used to evaluate the model’s ability to produce a plan that complies with practical standards. This is our core evaluation metric.


\singlespacing

\noindent\textbf{Baselines.} Our baselines are as follows:
\begin{itemize}[leftmargin=0.5cm]
    \item \noindent\textbf{Greedy Search}:
To evaluate the performance of traditional search algorithms in TravelPlanner, we adopt the greedy search strategy as one of the baselines. Greedy Search focuses on cost minimization as its core objective. Among transportation options, it selects the one with the lowest cost; for dining, it chooses restaurants with the lowest average expenditure; for accommodation, it selects the cheapest option; and for sightseeing, it arranges attractions by randomly selecting them each day. For a 5-day or 7-day travel plan, select the top 1 to 2 cities as destinations from the returned city search results.
\item  \noindent\textbf{Sole-Planning Mode}:
We focus on the sole-planning mode of the TravelPlan task. In this mode, the model is provided in advance with sufficient and necessary reference information required for reasoning and planning. This setting is used to evaluate the model's ability to perform complex reasoning and planning directly based on the given information. The baselines under the sole-planning mode include the following models and strategies:
\begin{itemize}[leftmargin=0.35cm]
    \item \textit{Models:} 
    GPT-3.5-Turbo, GPT-4-Turbo, GPT-4o (version: 2024-11-20), Mixtral-8×7B-MoE, Gemini Pro, Qwen3-8B-Instruct, DeepSeek-R1, our self-built Multi-Agent System mentioned above, and the qwen3-8b-instruct model trained only with Agentic SFT mentioned above.
    
    \item \textit{Strategies}: Our baselines include the following strategies: Direct, CoT, ReAct, Reflexion, and prompt reflect. Specifically, 
"Direct" refers to prompting the model to directly generate the final travel plan;
"CoT" refers to prompting the model to reason step by step before producing the final travel plan;
"ReAct" refers to prompting the model to solve the task by alternating between Thought, Action, and Observation steps;
"Reflexion" refers to prompting the model to perform self-reflection before generating the final travel plan;
"Prompt reflect" refers to prompting the model to reflect on its previous reasoning before producing the final travel plan.
\end{itemize}
    

\end{itemize}





\begin{table*}[t]
\caption{ Experimental results comparing Imagine with different baselines on six criteria}
\makebox[\textwidth][c]{
\centering
\small
\setlength{\tabcolsep}{3.5pt}
\renewcommand{\arraystretch}{0.95}
\begin{tabular}{@{\hspace{0pt}}>{\raggedright\arraybackslash}p{3.2cm}cccccc@{}}
\toprule
\multirow{2}{*}{\textbf{Methods}} 
    & \multirow{2}{*}{\textbf{Delivery Rate}}
      & \multicolumn{2}{c}{\textbf{Commonsense Constraint}} 
        & \multicolumn{2}{c}{\textbf{Hard Constraint}} 
            & \multirow{2}{*}{\textbf{Final Pass Rate}} \\
 \cmidrule(lr){3-4} \cmidrule(lr){5-6}
    &  
      & Micro & Macro 
        & Micro & Macro 
            & \\
\midrule
$\text{Greedy Search}$           &100 &72.0 &0 &52.4 &31.8 &0 \\
$\text{GPT-3.5-Turbo}_\text{Direct}$           &100 &59.5 &2.7 &9.5 &4.4 &0.6 \\
$\text{GPT-3.5-Turbo}_\text{CoT} $          &100 &64.4 &2.3 &9.8 &3.8 &0.4 \\
$\text{GPT-3.5-Turbo}_\text{ReAct}$           &81.6 &45.9 &2.5 &10.7 &3.1 &0.7 \\
$\text{GPT-3.5-Turbo}_\text{Reflexion}$           &92.1 &52.1 &2.2 &9.9 &3.8 &0.6 \\
$\text{GPT-4-Turbo}_\text{Direct}$           &100 &80.6 &15.2 &44.3 &23.1 &4.4 \\
$\text{Mixtral-8x7B-MoE}_\text{Direct}$   &99.3 &67.0 &3.7 &3.9 &1.6 &0.7 \\
$\text{Gemini Pro}_\text{Direct}$           &93.7 &64.7 &7.9 &10.6 &4.7 &2.1 \\
$\text{GPT-4o-2024-11-20}_\text{CoT}$           &100 &84.175 &25.9 &49.869 &26.6 &7 \\
$\text{Qwen3-8B-Instruct}_\text{CoT}$           &100 &72.04 &10.7 &28 &21.8 &5.9 \\
$\text{Qwen3-8B-Instruct}_\text{prompt reflect}$           &100 &73.8375 &14.6 &27.9476 &24 &8.3 \\
$\text{DeepSeek-R1}_\text{CoT}$ &100 &92.65 &55.2 &74.15 &62 &40 \\
$\text{Multi-Agent System}$ &100 &93.6625 &60.3 &77.729 &67.8 &45.8 \\
$\text{Agentic SFT}_\text{qwen3-8b-instruct}$ &100 &93.425 &56.9 &75.546 &62.2 &38.6 \\
\rowcolor{sigmaBG}
\textbf{Imagine(Ours)}                    &\textbf{100} 
                                       &\textbf{99.0375}\uparrowtext{5.375} &\textbf{92.5}\uparrowtext{32.2} &\textbf{92.271}\uparrowtext{14.542} &\textbf{86.9}\uparrowtext{19.1} &\textbf{82.7}\uparrowtext{36.9} \\
\bottomrule
\end{tabular}
}
\label{tab:main-results}
\end{table*}

\noindent\textbf{Models.}
We adopt the Qwen3-8B-Instruct model as the base model for training.



\subsection{Main Results}
The main resutls are shown in Table \ref{tab:main-results}.
The experimental results clearly demonstrate that our proposed method, IMAGINE, is highly effective in enhancing the model’s capabilities in complex reasoning and planning. Specifically, compared to the strongest baseline we evaluated against (i.e., the Multi-Agent System), IMAGINE achieved significant improvements across all criteria. On the Commonsense Constraint Micro Pass Rate, IMAGINE reached a pass rate of 99.0375\%, which is nearly perfect and surpasses the strongest baseline by 5.375 percentage points; On the Commonsense Constraint Macro Pass Rate, IMAGINE achieved a pass rate of 92.5\%, outperforming the strongest baseline by a substantial margin of 32.2 percentage points; On the Hard Constraint Micro Pass Rate, IMAGINE achieved a pass rate of 92.271\%, exceeding the strongest baseline by 14.542 percentage points; On the Hard Constraint Macro Pass Rate, IMAGINE achieved a pass rate of 86.9\%, surpassing the strongest baseline by 19.1 percentage points; On the Final Pass Rate, IMAGINE achieved a pass rate of 82.7\%, significantly outperforming the strongest baseline by 36.9 percentage points.

In addition, our model is a single end-to-end small model with a model size of only 8B. While significantly outperforming all baseline models in terms of performance, it also demonstrates considerable advantages in inference efficiency and cost. This further highlights the practical value and strong competitiveness of the IMAGINE method.

\begin{figure}[t]
\centering
\subfigure[Model performance during Agentic SFT training]{\includegraphics[width=0.43\linewidth]{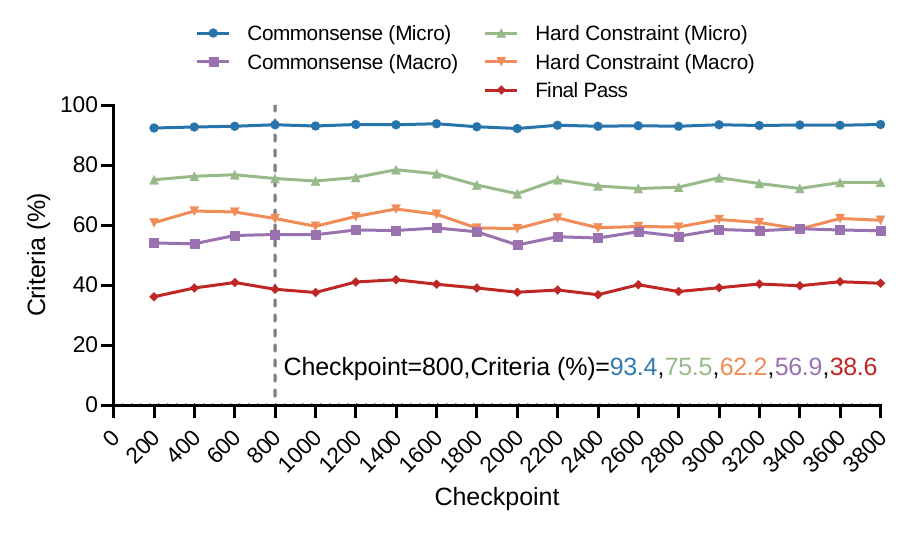}
\label{SFT}
}
\hspace{0.4cm}
\subfigure[Model Performance during Agentic GRPO training]{\includegraphics[width=0.47\linewidth]{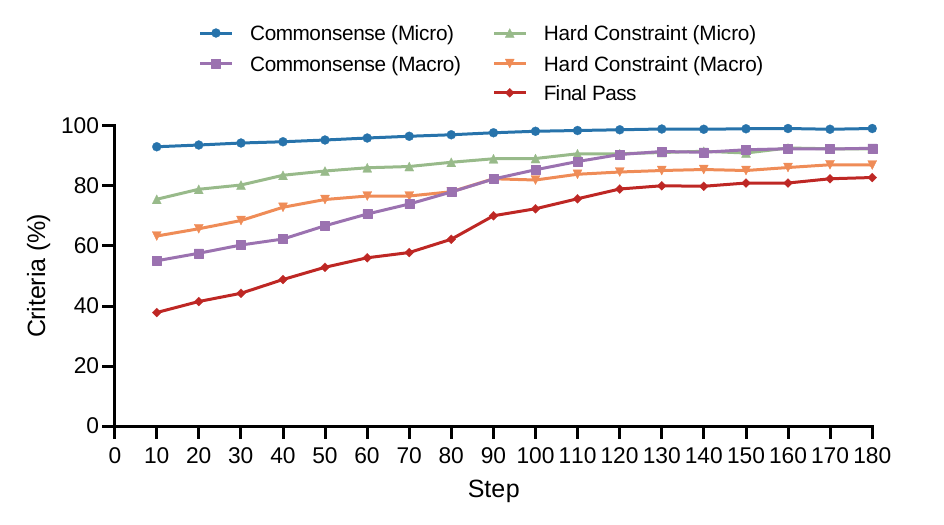}
\label{GRPO}
}
\caption{
The performance during Agentic SFT training and Agentic RL training
}
\end{figure}


\begin{figure}[t]
    \centering
    \includegraphics[width=1\linewidth]{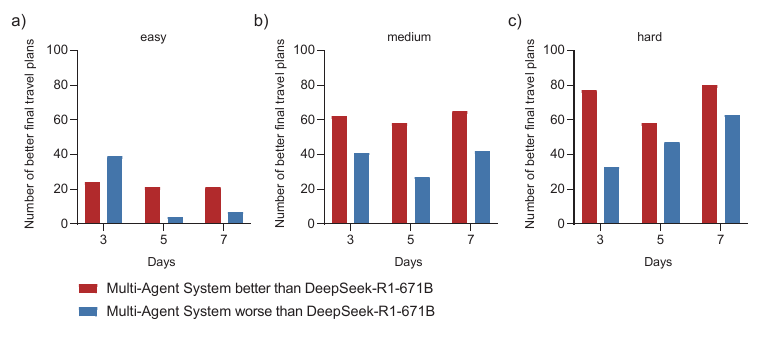}
    \caption{Comparison of the final travel plan produced}
    \label{Comparison_final_plan_MAS_DS-R1}
\end{figure}

\subsection{In-depth Analysis}
\begin{figure}[t]
    \centering
    \includegraphics[width=0.6\linewidth]{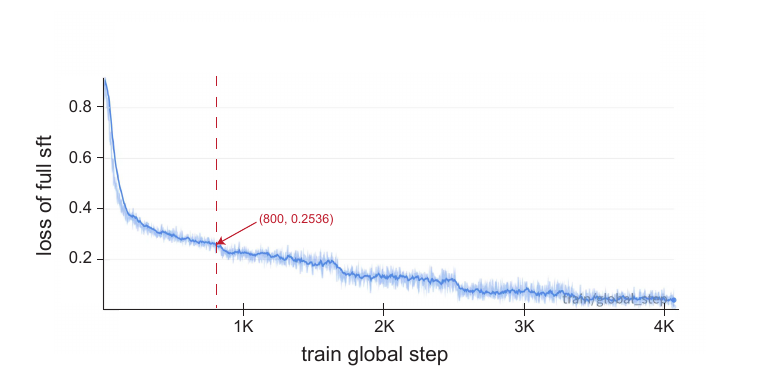}
    \caption{The loss during Agentic SFT training}
    \label{full_sft_loss}
\end{figure}

\paragraph{Analysis of SFT and GRPO.}
%
Considering that Supervised Fine-Tuning (SFT) serves as the warm-up stage before Reinforcement Learning (RL), the model trained during SFT needs to possess both a certain level of exploration capability and instruction-following ability to facilitate effective RL training. Therefore, the SFT stage should be neither under-trained nor over-trained.
If the model is under-trained during SFT, it will not have developed a basic alignment with the training data, making it difficult to obtain positive reward signals in the subsequent RL stage, and thus hindering the overall training effectiveness. Conversely, if the model is over-trained during SFT, its internal probability distribution may gradually collapse into one or a few fixed patterns. This results in the loss of sufficient exploration space during the RL stage, making it difficult for the model to discover more optimal actions—an outcome that is highly detrimental to reinforcement learning.
Figure \ref{full_sft_loss} shows the loss curve during our Agentic SFT training of the Qwen3-8B-Instruct model. Based on the consideration above, we selected the checkpoint at step 800 (checkpoint 800) during the intermediate phase of Agentic SFT training as the starting model for our Agentic RL training. 



\singlespacing

\noindent\textbf{Model performance during Agentic SFT training.}
Utilizing the inference data generated by the Multi-Agent System as described in the previous section on "Multi-Agent System based Inference Data Generation", we performed full SFT on the Qwen-3-8B-Instruct model using eight H800 GPUs (each with 80GB memory), with the learning rate set to 1.0e-5. Figure \ref{SFT} presents the test results on the TravelPlanner test dataset (1,000 queries) for every 200 checkpoints during the full SFT training process.
%





\singlespacing

\noindent\textbf{Model Performance during Agentic GRPO training.}
We continued reinforcement learning training on the checkpoint at step 800 (checkpoint 800), which had been previously trained using the Agentic SFT approach, by applying the GRPO algorithm. In this stage, we used our custom-designed rule-based reward function as the reward signal. The training data for the Agentic RL stage consisted of 225 queries in total, comprising the training set (45 queries) and validation set (180 queries) from the original TravelPlanner dataset. The training was conducted on 8 H800 GPUs (each with 80GB of memory), with the learning rate set to 1.0e-6.
Figure \ref{GRPO} shows the evaluation results on the TravelPlanner test dataset (1,000 queries), conducted every 10 steps during the reinforcement learning process. As illustrated in the figure, during Agentic GRPO training, the model demonstrated consistent improvements across all evaluation metrics, including Commonsense Constraint Micro Pass Rate, Commonsense Constraint Macro Pass Rate, Hard Constraint Micro Pass Rate, Hard Constraint Macro Pass Rate, and Final Pass Rate.

%




\singlespacing

\noindent\textbf{Comparison of the final travel plan.}
To verify whether the final travel plan generated by our designed Multi-Agent System outperforms that generated by the standalone DeepSeek-R1-671B model, we conducted a rigorous comparison based on format compliance, commonsense constraint satisfaction, and hard constraint satisfaction. The detailed results are shown in Figure \ref{Comparison_final_plan_MAS_DS-R1}.
In the figure, the red bars represent the number of cases where the final travel plan generated by the Multi-Agent System is strictly superior to that generated by the standalone DeepSeek-R1-671B model, while the blue bars indicate the number of cases where it is strictly inferior. As observed from the figure, except for the 3day easy dataset, the final travel plan produced by the Multi-Agent System generally outperforms that of the standalone DeepSeek-R1-671B model. This demonstrates the effectiveness of our designed Multi-Agent System for this task.

\begin{figure}[t]
    \centering
    \includegraphics[width=0.5\linewidth]{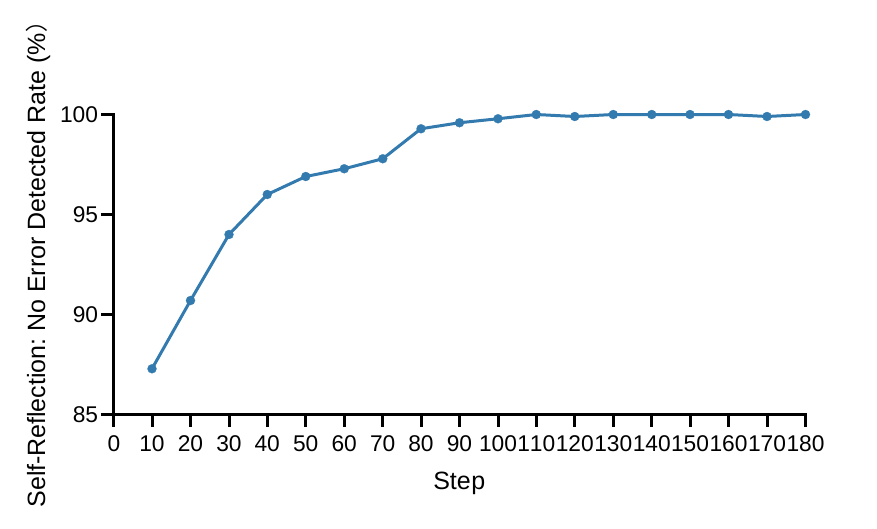}
    \caption{Proportion of "no error" in reflections}
    \label{Self-Reflection_No_Error_Detected}
\end{figure}

\singlespacing

\noindent\textbf{Reflection Analysis.}
We also evaluated the proportion of instances where the model determined that no errors existed in its previous reasoning during the Agentic GRPO training process after Agentic SFT, assessed at every 10 steps. Our evaluation was conducted on the original TravelPlanner test set, which consists of 1,000 queries. The specific results are shown in Figure \ref{Self-Reflection_No_Error_Detected}.
Combining Figure \ref{Self-Reflection_No_Error_Detected} and Figure \ref{GRPO}, it can be observed that as GRPO training progresses, the model increasingly tends to believe that there are no errors in its previous reasoning during the reflection phase, and the quality of the final travel plan generated by the model also gradually improves.

\section{Conclusion}
In this work, we propose IMAGINE, a novel and general framework designed to not only integrate the sophisticated reasoning and planning capabilities of Multi-Agent System into a single, compact language model but also to significantly surpass the capabilities of the original Multi-Agent System. Our method significantly enhances the performance of a single language model on complex reasoning and planning tasks, while completely avoiding the inherent limitations of Multi-Agent Systems (such as high inference costs and significant latency due to multi-turn interactions, and difficulties in end-to-end training).

Extensive experiments on the highly challenging TravelPlanner benchmark demonstrate the effectiveness of our approach. After training with our method, our 8B model achieves an 82.7\% Final Pass Rate on the TravelPlanner test set (1000 queries), significantly outperforming models such as GPT-4o (7\%) and DeepSeek-R1-671B (40\%), and surpassing our carefully designed Multi-Agent System (45.8\%) by a notable margin. Furthermore, the model exhibits substantial improvements across all evaluation criteria.

In addition to its exceptional performance, IMAGINE offers substantial practical advantages. The resulting single model operates with high efficiency, eliminating the need for multi-turn internal interactions or complex procedural workflows. This leads to a significant reduction in inference latency, inference cost, and deployment overhead. Its compact model size, combined with end-to-end trainability, ensures high scalability and ease of deployment. This study establishes a new paradigm for developing powerful and efficient reasoning models, demonstrating that a single model trained with our method can indeed learn and ultimately surpass the capabilities of a well-organized multi-agent team.


\medskip


\begin{thebibliography}{10}\itemsep=-1pt

\bibitem{chen2024survey}
S.~Chen, Y.~Liu, W.~Han, W.~Zhang, and T.~Liu.
\newblock A survey on llm-based multi-agent system: Recent advances and new frontiers in application.
\newblock {\em arXiv preprint arXiv:2412.17481}, 2024.

\bibitem{chen2024reprompt}
W.~Chen, S.~Koenig, and B.~Dilkina.
\newblock Reprompt: Planning by automatic prompt engineering for large language models agents.
\newblock {\em arXiv preprint arXiv:2406.11132}, 2024.

\bibitem{cheng2025revisiting}
Z.~Cheng, S.~Hao, T.~Liu, F.~Zhou, Y.~Xie, F.~Yao, Y.~Bian, Y.~Zhuang, N.~Dey, Y.~Zha, et~al.
\newblock Revisiting reinforcement learning for llm reasoning from a cross-domain perspective.
\newblock {\em arXiv preprint arXiv:2506.14965}, 2025.

\bibitem{choi2025atlas}
J.~Choi, J.~Yoon, J.~Chen, S.~Jha, and T.~Pfister.
\newblock Atlas: Constraints-aware multi-agent collaboration for real-world travel planning.
\newblock {\em arXiv preprint arXiv:2509.25586}, 2025.

\bibitem{de2008z3}
L.~De~Moura and N.~Bj{\o}rner.
\newblock Z3: An efficient smt solver.
\newblock In {\em International conference on Tools and Algorithms for the Construction and Analysis of Systems}, pages 337--340. Springer, 2008.

\bibitem{gandhi2025cognitive}
K.~Gandhi, A.~Chakravarthy, A.~Singh, N.~Lile, and N.~D. Goodman.
\newblock Cognitive behaviors that enable self-improving reasoners, or, four habits of highly effective stars.
\newblock {\em arXiv preprint arXiv:2503.01307}, 2025.

\bibitem{guo2025deepseek}
D.~Guo, D.~Yang, H.~Zhang, J.~Song, R.~Zhang, R.~Xu, Q.~Zhu, S.~Ma, P.~Wang, X.~Bi, et~al.
\newblock Deepseek-r1: Incentivizing reasoning capability in llms via reinforcement learning.
\newblock {\em arXiv preprint arXiv:2501.12948}, 2025.

\bibitem{guo2024large}
T.~Guo, X.~Chen, Y.~Wang, R.~Chang, S.~Pei, N.~V. Chawla, O.~Wiest, and X.~Zhang.
\newblock Large language model based multi-agents: A survey of progress and challenges.
\newblock {\em arXiv preprint arXiv:2402.01680}, 2024.

\bibitem{guo2025mirror}
Z.~Guo, B.~Xu, X.~Wang, and Z.~Mao.
\newblock Mirror: Multi-agent intra-and inter-reflection for optimized reasoning in tool learning.
\newblock {\em arXiv preprint arXiv:2505.20670}, 2025.

\bibitem{hao2024large}
Y.~Hao, Y.~Chen, Y.~Zhang, and C.~Fan.
\newblock Large language models can solve real-world planning rigorously with formal verification tools.
\newblock {\em arXiv preprint arXiv:2404.11891}, 2024.

\bibitem{hong2024metagpt}
S.~Hong, M.~Zhuge, J.~Chen, X.~Zheng, Y.~Cheng, C.~Zhang, J.~Wang, Z.~Wang, S.~K.~S. Yau, Z.~Lin, et~al.
\newblock Metagpt: Meta programming for a multi-agent collaborative framework.
\newblock International Conference on Learning Representations, ICLR, 2024.

\bibitem{jaech2024openai}
A.~Jaech, A.~Kalai, A.~Lerer, A.~Richardson, A.~El-Kishky, A.~Low, A.~Helyar, A.~Madry, A.~Beutel, A.~Carney, et~al.
\newblock Openai o1 system card.
\newblock {\em arXiv preprint arXiv:2412.16720}, 2024.

\bibitem{li2023camel}
G.~Li, H.~Hammoud, H.~Itani, D.~Khizbullin, and B.~Ghanem.
\newblock Camel: Communicative agents for" mind" exploration of large language model society.
\newblock {\em Advances in Neural Information Processing Systems}, 36:51991--52008, 2023.

\bibitem{liu2025symagent}
B.~Liu, J.~Zhang, F.~Lin, C.~Yang, M.~Peng, and W.~Yin.
\newblock Symagent: A neural-symbolic self-learning agent framework for complex reasoning over knowledge graphs.
\newblock In {\em Proceedings of the ACM on Web Conference 2025}, pages 98--108, 2025.

\bibitem{liu2025understanding}
Z.~Liu, C.~Chen, W.~Li, P.~Qi, T.~Pang, C.~Du, W.~S. Lee, and M.~Lin.
\newblock Understanding r1-zero-like training: A critical perspective.
\newblock {\em arXiv preprint arXiv:2503.20783}, 2025.

\bibitem{ou2025analyzing}
T.~Ou, S.~Vaduguru, and D.~Fried.
\newblock Analyzing information sharing and coordination in multi-agent planning.
\newblock {\em arXiv preprint arXiv:2508.12981}, 2025.

\bibitem{qian2025scaling}
C.~Qian, Z.~Xie, Y.~Wang, W.~Liu, K.~Zhu, H.~Xia, Y.~Dang, Z.~Du, W.~Chen, C.~Yang, Z.~Liu, and M.~Sun.
\newblock Scaling large language model-based multi-agent collaboration.
\newblock In {\em The Thirteenth International Conference on Learning Representations}, 2025.

\bibitem{qiao2025thematic}
T.~Qiao, C.~Walker, C.~Cunningham, and Y.~S. Koh.
\newblock Thematic-lm: a llm-based multi-agent system for large-scale thematic analysis.
\newblock In {\em Proceedings of the ACM on Web Conference 2025}, pages 649--658, 2025.

\bibitem{schulman2017proximal}
J.~Schulman, F.~Wolski, P.~Dhariwal, A.~Radford, and O.~Klimov.
\newblock Proximal policy optimization algorithms.
\newblock {\em arXiv preprint arXiv:1707.06347}, 2017.

\bibitem{shao2024deepseekmath}
Z.~Shao, P.~Wang, Q.~Zhu, R.~Xu, J.~Song, X.~Bi, H.~Zhang, M.~Zhang, Y.~Li, Y.~Wu, et~al.
\newblock Deepseekmath: Pushing the limits of mathematical reasoning in open language models.
\newblock {\em arXiv preprint arXiv:2402.03300}, 2024.

\bibitem{shinn2023reflexion}
N.~Shinn, F.~Cassano, A.~Gopinath, K.~Narasimhan, and S.~Yao.
\newblock Reflexion: Language agents with verbal reinforcement learning.
\newblock {\em Advances in Neural Information Processing Systems}, 36:8634--8652, 2023.

\bibitem{wang2025cooperative}
Z.~Wang, Z.~Zhao, Y.~Lyu, Z.~Chen, M.~de~Rijke, and Z.~Ren.
\newblock A cooperative multi-agent framework for zero-shot named entity recognition.
\newblock In {\em Proceedings of the ACM on Web Conference 2025}, pages 4183--4195, 2025.

\bibitem{wei2022chain}
J.~Wei, X.~Wang, D.~Schuurmans, M.~Bosma, F.~Xia, E.~Chi, Q.~V. Le, D.~Zhou, et~al.
\newblock Chain-of-thought prompting elicits reasoning in large language models.
\newblock {\em Advances in neural information processing systems}, 35:24824--24837, 2022.

\bibitem{10.1007/BF00992696}
R.~J. Williams.
\newblock Simple statistical gradient-following algorithms for connectionist reinforcement learning.
\newblock {\em Mach. Learn.}, 8(3–4):229–256, May 1992.

\bibitem{xie2024travelplanner}
J.~Xie, K.~Zhang, J.~Chen, T.~Zhu, R.~Lou, Y.~Tian, Y.~Xiao, and Y.~Su.
\newblock Travelplanner: A benchmark for real-world planning with language agents.
\newblock {\em arXiv preprint arXiv:2402.01622}, 2024.

\bibitem{xiong2025minimalist}
W.~Xiong, J.~Yao, Y.~Xu, B.~Pang, L.~Wang, D.~Sahoo, J.~Li, N.~Jiang, T.~Zhang, C.~Xiong, et~al.
\newblock A minimalist approach to llm reasoning: from rejection sampling to reinforce.
\newblock {\em arXiv preprint arXiv:2504.11343}, 2025.

\bibitem{yang2025plan}
D.~Yang, C.~Lu, Q.~Wang, X.~Ma, Y.~Gao, Y.~Hu, and H.~Zhao.
\newblock Plan your travel and travel with your plan: Wide-horizon planning and evaluation via llm.
\newblock {\em arXiv preprint arXiv:2506.12421}, 2025.

\bibitem{yu2025dapo}
Q.~Yu, Z.~Zhang, R.~Zhu, Y.~Yuan, X.~Zuo, Y.~Yue, W.~Dai, T.~Fan, G.~Liu, L.~Liu, et~al.
\newblock Dapo: An open-source llm reinforcement learning system at scale.
\newblock {\em arXiv preprint arXiv:2503.14476}, 2025.

\bibitem{yue2025does}
Y.~Yue, Z.~Chen, R.~Lu, A.~Zhao, Z.~Wang, S.~Song, and G.~Huang.
\newblock Does reinforcement learning really incentivize reasoning capacity in llms beyond the base model?
\newblock {\em arXiv preprint arXiv:2504.13837}, 2025.

\bibitem{zeeshan2025large}
T.~Zeeshan, A.~Kumar, S.~Pirttikangas, and S.~Tarkoma.
\newblock Large language model based multi-agent system augmented complex event processing pipeline for internet of multimedia things.
\newblock {\em arXiv preprint arXiv:2501.00906}, 2025.

\bibitem{zhang2024planning}
C.~Zhang, D.~G.~X. Deik, D.~Li, H.~Zhang, and Y.~Liu.
\newblock Planning with multi-constraints via collaborative language agents.
\newblock {\em arXiv preprint arXiv:2405.16510}, 2024.

\bibitem{zhang2025critique}
X.~Zhang, H.~Sun, Y.~Zhang, K.~Feng, C.~Lu, C.~Yang, and H.~Meng.
\newblock Critique-grpo: Advancing llm reasoning with natural language and numerical feedback.
\newblock {\em arXiv preprint arXiv:2506.03106}, 2025.

\bibitem{zhang2025swarmagentic}
Y.~Zhang, C.~Lin, S.~Tang, H.~Chen, S.~Zhou, Y.~Ma, and V.~Tresp.
\newblock Swarmagentic: Towards fully automated agentic system generation via swarm intelligence.
\newblock {\em arXiv preprint arXiv:2506.15672}, 2025.

\end{thebibliography}

\end{document}